\documentclass[letterpaper, 10 pt, conference]{ieeeconf}
\usepackage[T1]{fontenc}
\IEEEoverridecommandlockouts                                  
\usepackage{times}
\usepackage{epsfig}
\usepackage{graphicx}
\usepackage{amsmath}
\usepackage{amssymb}
\usepackage{multirow}
\usepackage{comment}
\usepackage{color}
\usepackage{multirow}
\usepackage{makecell}
\usepackage{times}
\usepackage{bbding}
\usepackage{pifont}
\usepackage{float}
\usepackage{cite}
\usepackage{threeparttable}

\usepackage[hidelinks,colorlinks=true,linkcolor=blue,citecolor=blue,urlcolor=blue]{hyperref}
\usepackage{booktabs}
\usepackage[linesnumbered,ruled,vlined]{algorithm2e} 
\SetKwInput{KwIn}{Input}    
\SetKwInput{KwOut}{Output}

\title{\LARGE \bf
CoordField: Coordination Field for Agentic UAV Task Allocation \\ In Low-altitude Urban Scenarios}

\author{
Tengchao Zhang$^{1}$ , Yonglin Tian$^{2}$ $^{\dagger}$, Fei Lin$^{1}$, Jun Huang$^{1}$,Patrik P. Süli$^{3}$, Qinghua Ni$^{1}$, Rui Qin$^{4}$, \\Xiao Wang$^{5}$, and Fei-Yue Wang$^{6}$  
\thanks{$^{\dagger}$Corresponding author: Yonglin Tian (email: yonglin.tian@ia.ac.cn).}
\thanks{This work was partly supported by the Science and Technology Development Fund, Macau Special Administrative Region (SAR) (0145/2023/RIA3, 0093/2023/RIA2, 0157/2024/RIA2), the Project supported by Laboratory of Computation and Analytics of Complex Management Systems (Tianjin University), and National Natural Science Foundation of China (\#72171230); this is part of the LaSEE and CiSEE projects.} 
\thanks{$^{1}$Tengchao Zhang, Fei Lin, Jun Huang and Qinghua Ni are with the Department of Engineering Science, Faculty of Innovation Engineering, Macau University of Science and Technology, Macau 999078, China (e-mail: zhangtengchao@ieee.org, feilin@ieee.org, junhuang@ieee.org, qinghua.ni@ieee.org)}
\thanks{$^{2}$Yonglin Tian is with the State Key Laboratory of Multimodal Artificial Intelligence Systems, Institute of Automation, Chinese Academy of Sciences, Beijing 100190, China (e-mail: yonglin.tian@ia.ac.cn).}
\thanks{$^{3}$Patrik P. Süli is with the Doctoral School of Applied Informatics and Applied Mathematics, Óbuda University, Hungary, (e-mail: suli.patrik@nik.uni-obuda.hu).}
\thanks{$^{4}$Rui Qin is with the State Key Laboratory of Multimodal Artificial Intelligence Systems, Institute of Automation, Chinese Academy of Sciences, Beijing 100190, China, and also with Laboratory of Computation and Analytics of Complex Management Systems, Tianjin University, Tianjin 300072, China (e-mail: rui.qin@ia.ac.cn).}
\thanks{$^{5}$Xiao Wang is with the School of Artificial Intelligence, Anhui University, Anhui 230601, China. (e-mail: xiao.wang@ahu.edu.cn).}
\thanks{$^{6}$Fei-Yue Wang is with the State Key Laboratory for Management and Control of Complex Systems, Chinese Academy of Sciences, Beijing 100190, and also with the Department of Engineering Science, Faculty of Innovation Engineering, Macau University of Science and Technology, Macau 999078, China, and also with University Research and Innovation Center, Obuda University, Budapest, Hungary. (e-mail: feiyue.wang@ia.ac.cn)}%
}

\usepackage{float}
\begin{document}

\maketitle
\thispagestyle{empty}
\pagestyle{empty}


\begin{abstract}
With the increasing demand for heterogeneous Unmanned Aerial Vehicle (UAV) swarms to perform complex tasks in urban environments, system design now faces major challenges, including efficient semantic understanding, flexible task planning, and the ability to dynamically adjust coordination strategies in response to evolving environmental conditions and continuously changing task requirements. To address the limitations of existing methods, this paper proposes CoordField, a coordination field agent system for coordinating heterogeneous drone swarms in complex urban scenarios. In this system, large language models (LLMs) is responsible for interpreting high-level human instructions and converting them into executable commands for the UAV swarms, such as patrol and target tracking. Subsequently, a Coordination field mechanism is proposed to guide UAV motion and task selection, enabling decentralized and adaptive allocation of emergent tasks. A total of 50 rounds of comparative testing were conducted across different models in a 2D simulation space to evaluate their performance. Experimental results demonstrate that the proposed system achieves superior performance in terms of task coverage, response time, and adaptability to dynamic changes.
\end{abstract}


\section{Introduction}

The growing demand for deploying Unmanned Aerial Vehicle (UAV) swarms to perform tasks such as pedestrian detection, vehicle tracking, and traffic signal monitoring in complex and dynamic urban environments has highlighted the advantages of multi-UAV systems over single UAV, particularly in terms of coverage, system redundancy, and operational efficiency. However, the increasing complexity of tasks and the highly dynamic nature of urban environments also pose significant challenges\cite{makridis2024real}. This requires effective task allocation methods for drone swarms to adapt to ever-changing new tasks, while also achieving real-time intelligent scheduling of the system.

To address these challenges, various optimization algorithms have been developed for UAV swarm coordination, such as the Grey Wolf Optimizer \cite{yang2025hybrid} and the Whale Optimization Algorithm \cite{xu2025dbo}. However, these methods are typically tailored to specific types of tasks and struggle to adapt to heterogeneous UAV clusters operating in highly dynamic environments \cite{selvaraj2023learning}. With the rise of large language models (LLMs), recent research has explored leveraging their powerful semantic understanding and tool-calling capabilities to address such problems. For example, several approaches attempt to utilize few-shot or zero-shot learning methods based on existing models, enabling LLMs to perform multi-round planning, itinerary calibration, and execution \cite{song2023llm,huang2023instruct2act}. Nevertheless, such methods still exhibit significant limitations when handling multiple tasks in complex and rapidly changing environments.

To address the above challenges, the emerging Agentic Artificial Intelligence(AI) offers a more flexible and unified approach \cite{acharya2025agentic}. This paradigm is driven by LLMs, and integrates core functionalities such as task decomposition, tool invocation, and multi-agent coordination and scheduling into a cohesive system \cite{qiu2024llm,xu2025mem}. In this paper, we propose a coordination field multi-UAV agentic system architecture. The system leverages LLM to parse natural language instructions into executable tasks for UAV swarms, and employs a coordination field control strategy to achieve task-oriented autonomous navigation and collective coordination.

The main contributions of this work are as follows:
\begin{itemize}
    \item We propose CoordField, a coordination field-based UAV swarms task allocation method that employs continuously updated potential fields to represent both task urgency and UAV influence within urban environments. This approach enhances real-time responsiveness.
    \item We present an agentic system for UAV swarms that enables high-precision task understanding from natural language descriptions and supports dynamic planning and deployment through specialized collaborative agents with distinct roles.
    \item We evaluate our system against multiple baseline models under the same task scenarios, validating its superior coordination performance in urban environments.
\end{itemize}

\section{Preliminaries}

\subsection{Agentic Systems}

In recent years, Agentic AI has emerged as a novel intelligent paradigm characterized by autonomous perception, goal-driven planning, and adaptive decision-making, offering innovative and effective solutions for multi-agent collaborative tasks \cite{chen2023agentverse,tran2025multi}. Unlike traditional passive approaches, Agentic AI emphasizes continuous goal pursuit, contextual memory, adaptive tool use, and self-reflection capabilities of agents. These abilities enable agents to operate with greater autonomy and coordination in dynamic environments. Agent systems powered by LLMs have demonstrated strong reasoning and planning capabilities, particularly in complex task decomposition and sequential execution \cite{song2023llm}. These features provide a solid theoretical and methodological foundation for building autonomous multi-UAV systems. In complex urban task scenarios \cite{tian2025uavs}, Agentic AI can assist UAVs in understanding high-level instructions, distributing subtasks effectively, and rapidly reconfiguring coordination strategies in response to task changes or agent failures, thereby enabling truly autonomous collaboration \cite{lin2024airvista, lin2025airvista}.

\subsection{LLMs for Natural Language Task Parsing}

With the rapid advancement of LLMs in natural language understanding and general reasoning, LLMs have demonstrated superior capabilities over traditional rule-based parsers in handling ambiguous semantics, conditional logic, and spatial language\cite{cui2024receive}. They also exhibit strong generalization in task generation, even under limited domain-specific data. Researchers have developed several frameworks such as Code-as-Policies and Prompt2Action\cite{liang2023code,liang2024taskmatrix}, enabling LLMs to map natural language instructions into task-executable codes. These approaches often incorporate few-shot prompting and constraint prompting to improve the validity and safety of the generated outputs. Furthermore, the GSCE framework, which consists of Guidelines, Skill APIs, Constraints, and Examples, has demonstrated the practical value of prompt engineering in UAV task planning\cite{wang2025gsce}, guiding LLMs to generate physically feasible and constraint-compliant control programs by providing skill APIs, operational constraints, and illustrative examples\cite{tian2024logisticsvista, lin2024securing}.

\subsection{Field-Based Coordination and Allocation}

Field-driven methods, particularly the Artificial Potential Field (APF) approach, have been widely applied in the field of mobile robotics\cite{fan2020improved}. In this approach, robots navigate by descending along the gradient of the resulting potential field, thereby enabling path planning and real-time obstacle avoidance. This approach has been applied applications in traffic flow modeling \cite{rostami2019obstacle}, where field-based methods are used to simulate the evolution of traffic dynamics. Vehicles are modeled as particles influenced by traffic density fields, velocity fields, and other factors, allowing the system to capture both macroscopic and microscopic variations more effectively .

Inspired by this line of thought, our work adopts a similar strategy by modeling task demand and UAV workload as continuous spatial fields. A dynamically evolving potential field is constructed based on the distribution of task urgency, guiding UAVs to move and allocate themselves through local gradient descent \cite{pan2021improved}. This method excels with its decentralized structure, real-time adaptability, and responsiveness to dynamic tasks, making it ideal for uneven and changing urban scenarios. Unlike traditional discrete optimization, field-driven approaches are more flexible and scalable.


\section{Approach}

\begin{figure*}
    \centering
    \includegraphics[width=\textwidth]{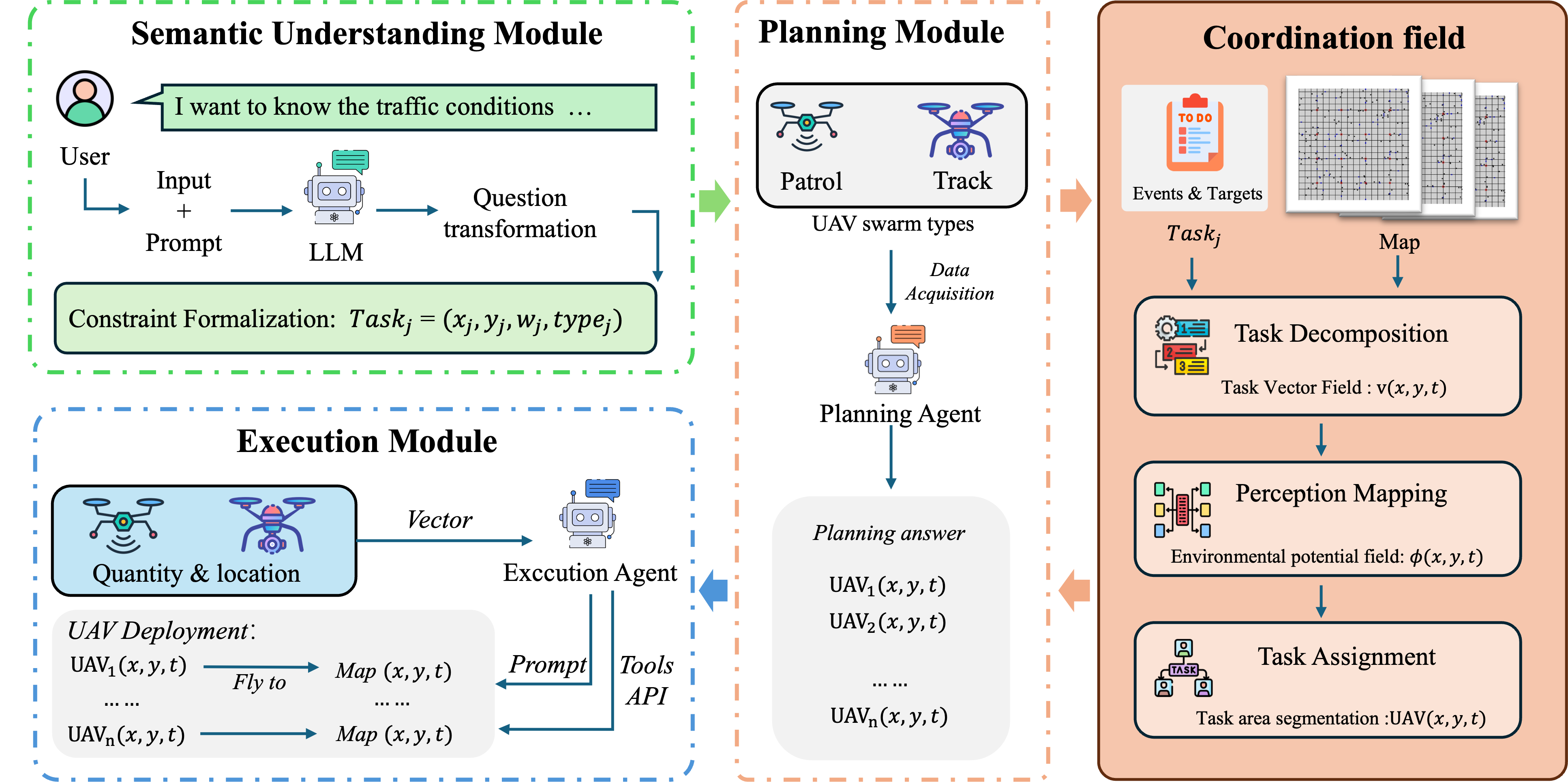}
    \caption{Coordination field Agentic System Implementation Process}
    \label{fig:1}
\end{figure*}

In this work, we propose an Agentic system for UAV swarms, built upon a Coordination field mechanism is proposed to guide UAV motion and task selection, enabling decentralized and adaptive allocation of emergent tasks. methodology. The proposed agentic system consists of three primary modules, as illustrated in Fig. \ref{fig:1}. It enables multi-layered agent coordination to interpret natural language inputs and perform task decomposition and allocation for UAV swarms in urban environments. In particular, to address the heterogeneity of UAV types and the complexity of urban traffic conditions, a field-driven strategy is adopted to guide different types of UAVs toward task regions that match their operational roles.

\subsection{Semantic Understanding Module}

The semantic understanding module is responsible for interpreting user-provided natural language inputs and converting them into structured task representations suitable for downstream planning and execution. Serving as the human–system interface, it enables intuitive interaction with the UAV swarm through language-based commands and ensures that high-level semantic information is accurately conveyed to subsequent modules in the system\cite{xu2024reasoning}.

Upon receiving a user prompt, the LLM performs a task translation process, mapping vague or abstract natural language instructions into formalized constraints. Specifically, the LLM processes the input to extract key elements such as target location, task type, and priority level, and ultimately generates a structured task tuple in the following form:

\begin{equation*}
\text{Task}_j = (x_j, y_j, w_j, \text{type}_j)
\end{equation*}

\noindent where $(x_j, y_j)$ denotes the spatial coordinates of the target, $w_j$ represents the task weight or priority, and $\text{type}_j$ refers to the semantic category of the task (e.g., patrol, tracking, inspection).

This formalization process acts as a bridge between high-level user intent and the low-level execution logic of the agentic system. It enables downstream modules, such as the planning and execution modules, to operate on a consistent and quantifiable task representation. Moreover, the module leverages the strong semantic understanding and reasoning capabilities of LLMs, laying a foundation for flexible task expansion and constraint integration.

\subsection{Planning Module and Coordination Field}

The Planning Module is responsible for transforming the task tuples $(x_j, y_j, w_j, \text{type}_j)$ obtained from semantic parsing into a dynamic UAV coordination strategy. The Planning Agent first feeds the task tuples and UAV state data into the coordination field. This field-based framework consists of three submodules—perception mapping, task decomposition, and task assignment—which operate in a closed-loop manner to continuously adapt to evolving task demands and spatial distribution patterns. The results are then aggregated and delivered by the Planning Agent for execution.

Perception Mapping: The Planning Agent first reads relevant map information and UAV configurations, such as types, capabilities, and quantities. When a task emerges at any given time $t$, the Planning Agent groups related tasks and identifies a set of active tasks, each $Task_j$ located at position $(x_j, y_j)$ with a dynamic urgency weight $w_j(t)$.

To establish task density awareness in the environment, we construct a time-varying potential field $\phi(x, y, t)$ to represent the spatial intensity and distribution of high-priority task regions. This scalar field is composed of a sum of weighted Gaussian functions, defined as:

\begin{equation*}
\phi(x,y,t) := \sum_{j=1}^{M} w_j(t) \cdot \exp\left(-\frac{\left\|(x,y) - (x_j(t), y_j(t))\right\|^2}{2\sigma_j^2}\right)
\end{equation*}

Here, $\sigma_j$ denotes the spatial influence radius of task $j$, reflecting the degree to which its urgency propagates across nearby areas. Notably, obstacles such as buildings are excluded from the field domain, and $\phi(x, y, t)$ is explicitly set to zero within these regions.

By constructing this continuous ``task landscape'', UAVs are able to perceive and exploit the gradient of $\phi(x, y, t)$ to identify and navigate toward high-priority task areas.

Task decomposition: To convert the scalar potential field $\phi$ into actionable guidance for UAV navigation and control, we further construct a time-varying vector field $\mathbf{v}(x, y, t)$, which represents the motion velocity field of the UAV swarm. During the process, the following motion equation is defined based on the Navier–Stokes equations in fluid dynamics:

\begin{equation*}
\frac{d\mathbf{v}}{dt} := -\frac{1}{\rho}\nabla p + \nu \nabla^2 \mathbf{v} + \mathbf{F}_{\text{task}}(\phi)
\end{equation*}

In this model, $\mathbf{v}(x, y, t) = (v_x, v_y)$ denotes the velocity vector at each point in the 2D space, indicating the direction and magnitude that a UAV should follow at a given position. The parameter $\nu$ represents the fluid viscosity, which governs the smoothness and responsiveness of the velocity field: higher viscosity results in smoother flows but slower response. $\mathbf{F}_{\text{task}}$ is the external force term used to steer UAVs toward regions with higher $\phi$ values, and it is defined as:

\begin{equation*}
\mathbf{F}_{\text{task}} = k \nabla \phi
\end{equation*}

\noindent where $k$ is a scaling coefficient that adjusts the attraction strength. $\rho$ is the nominal fluid density, which can be normalized in this system for simplification.

By constructing such a velocity field, UAVs do not need to explicitly plan trajectories. Instead, they can simply follow the local flow direction, naturally converging toward high-demand task regions. Meanwhile, the inclusion of the diffusion term $\nu \nabla^2 \mathbf{v}$ helps prevent over-concentration of UAVs in a single area, thereby enhancing spatial distribution and system robustness.

Task Assignment: To further enhance coordination efficiency among multiple agents and prevent resource redundancy or execution conflicts caused by UAVs clustering in the same task region, we introduce a local vortex mechanism around each UAV to serve as a dynamic repulsion control strategy. Specifically, each UAV $i$ generates a rotational field centered at its current position $(x_i, y_i)$, defined as:

\begin{equation*}
\omega_i(r) := \frac{\Gamma_i}{2\pi r} \exp\left(-\left(\frac{r}{r_0}\right)^2\right)
\end{equation*}

\noindent where $r$ is the radial distance from UAV $i$, and $r_0$ denotes the influence radius of the field. $\Gamma_i$ is the circulation strength generated by the UAV, which depends on its capability score $c_i(t)$ and the local potential value $\phi(x_i, y_i, t)$, calculated as:

\begin{equation*}
\Gamma_i(t) := \frac{c_i(t) \cdot \phi(x_i(t), y_i(t), t)}{\sum_{j=1}^{N} c_j(t)}
\end{equation*}

Based on this field, the system computes a tangential velocity component $v_{\theta,i}(r)$, which forms a repulsive velocity around each UAV and models the inter-agent avoidance dynamics:

\begin{equation*}
v_{\theta,i}(r) := \frac{\Gamma_i}{2\pi r} \left(1 - \exp\left(-\left(\frac{r}{r_0}\right)^2\right)\right)
\end{equation*}

Finally, the control velocity vector of each UAV is obtained by combining the global guiding velocity field $\mathbf{v}(x, y, t)$ with the superimposed tangential repulsive velocities generated by all UAVs:

\begin{equation*}
v_{\text{new}}(x, y, t) = v(x, y, t) + \sum_{i=1}^{n} v_{\theta,i}(x, y, t)
\end{equation*}

This mechanism ensures that UAVs are not only guided toward high-priority task regions but also mutually repelled through localized vortex effects, thereby enabling dynamically stable task allocation and coordination behavior in a fully decentralized setting. The system thus achieves high adaptability and strong scalability.

\subsection{Execution Module}
The core task of the Execution Module is to translate the outputs of the planning module, including the flow field vector $\mathbf{v}_{\text{new}}(x, y, t)$ and the UAV task assignment information, into real-time control commands executable by either physical or simulated UAVs. This process is managed by the Execution Agent, which communicates with the underlying control system through standardized API interfaces.

During the task execution phase, each UAV samples a local velocity vector from the precomputed control field $\mathbf{v}_{\text{new}}(x, y, t)$ based on its current position $(x, y, t)$, and uses this vector as its next navigation direction. The Execution Agent then generates flight control commands accordingly and dispatches them to the UAV platform (either physical or virtual) via middleware such as MAVSDK, ROS2, or AirSim. These control commands can be formalized as:

\begin{equation*}
\text{UAV}(x, y, t) \rightarrow \text{Map}(x^{\prime}, y^{\prime}, t^{\prime})
\end{equation*}

\noindent which maps the current state to a new target position and time step.

Based on the task type $\text{type}_j$, UAVs are assigned distinct execution modes, such as performing area patrol or tracking specific targets. These operational differences are embedded in the command structure during instruction generation and are reflected in the API parameters.

To maintain adaptability in dynamic environments, each UAV continuously reports its status, including position, velocity, and execution state, to the control system. These updates are used to refresh the task potential field $\phi(x, y, t)$ and the flow field vector $\mathbf{v}(x, y, t)$ in real time. When the environment changes, such as when new task targets appear, current tasks are completed, or obstacles are introduced, the system recalculates the relevant fields and delivers updated control instructions through the Execution Agent.

This closed-loop perception–planning–execution cycle operates at a high frequency, ensuring the system exhibits strong responsiveness and decentralized dynamic coordination. Under this mechanism, UAVs are capable of re-planning, cooperative obstacle avoidance, and target adaptation in complex and ever-changing environments, thereby realizing truly adaptive intelligent behavior.


\section{Experiments}

\begin{figure}[!htbp]
    \centering
    \includegraphics[width=1.0\columnwidth]{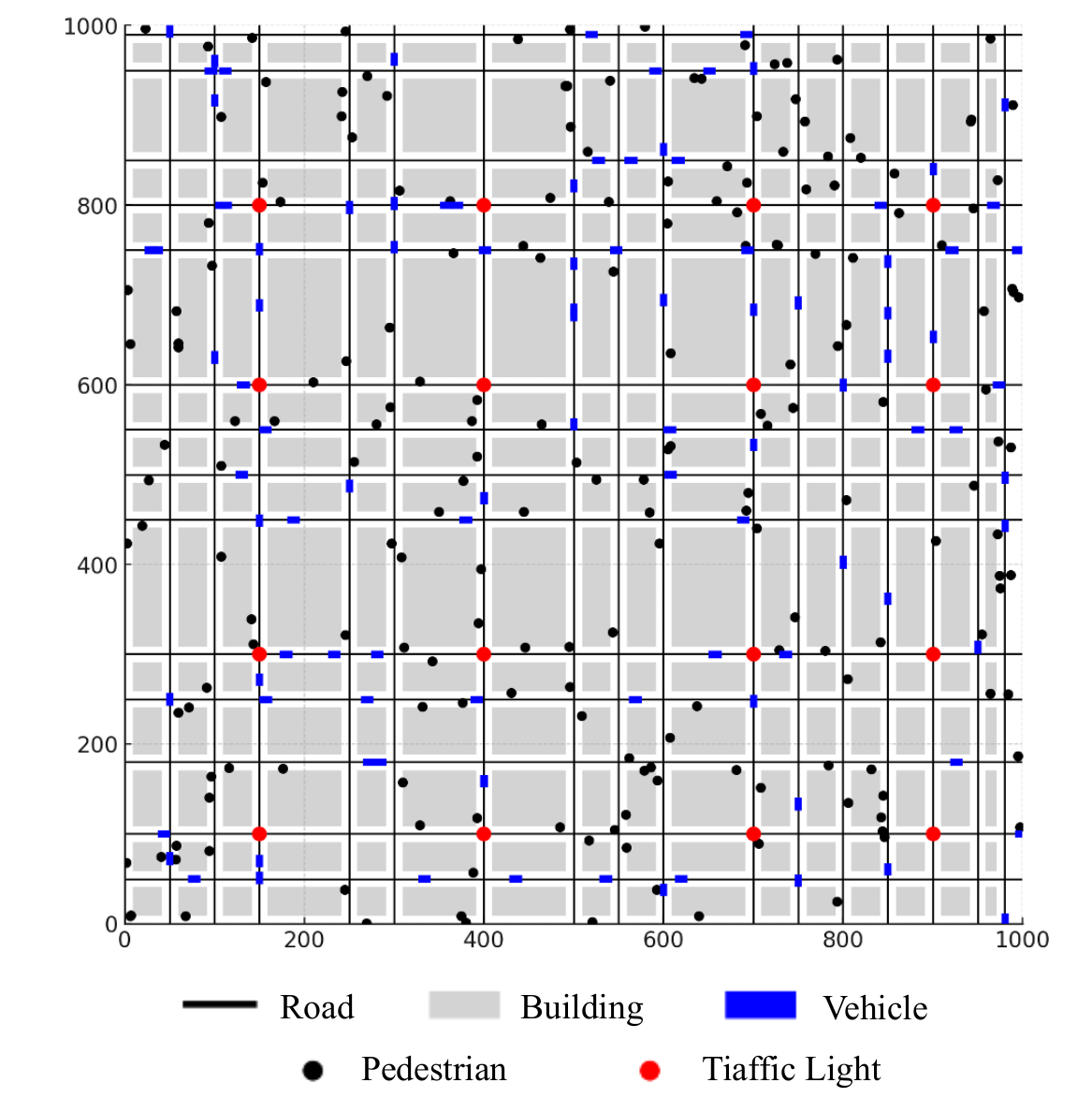}
    \caption{2D space simulation map.}
    \label{fig:2}
\end{figure}

To validate the proposed architecture, we conducted a series of experiments in a simulated urban environment. We first introduce the construction of the simulation environment, the UAV swarm configuration, the task generation process—including natural language instruction parsing—and the baseline methods used for comparison. We then describe the performance evaluation metrics and testing procedures. Finally, we present results from various experimental scenarios to demonstrate the system’s effectiveness in terms of task coverage, load balancing, and response latency.

\subsection{Environment and UAV Configuration}

The study is conducted within a custom-designed two-dimensional urban simulation environment, covering a $1000 \times  1000$ grid space, as illustrated in Fig. \ref{fig:2}. The environment features a structured city road network, with gray regions representing densely distributed buildings. Red dots mark traffic lights located at road intersections, while a large number of black points and blue rectangles denote dynamically moving pedestrians and vehicles, respectively. The roads are laid out in alignment with the city grid, creating a complex traffic and pedestrian flow environment for multi-agent interaction and coordination.

In the simulation, 20 UAVs operate at a fixed, safe altitude above a $1000 \times 1000$ urban road network. They are evenly divided into patrol and tracking types—sharing identical flight dynamics but differing in sensing capabilities, with patrol UAVs suited for wide-area scanning and tracking UAVs for precise target following. Each UAV is assigned a dynamic capability vector $\mathbf{c}_i(t)$ to reflect task adaptability. The environment updates in real time with step $T$, capturing pedestrian and vehicle dynamics and traffic signal changes. Task commands, such as “Please inspect the crowd and vehicles,” are issued via natural language and parsed by the DeepSeek API into structured task parameters including location, type, and priority.

\subsection{Experimental Results and Analysis}

\begin{figure}[!htbp]
    \centering
    \includegraphics[width=1.0\columnwidth]{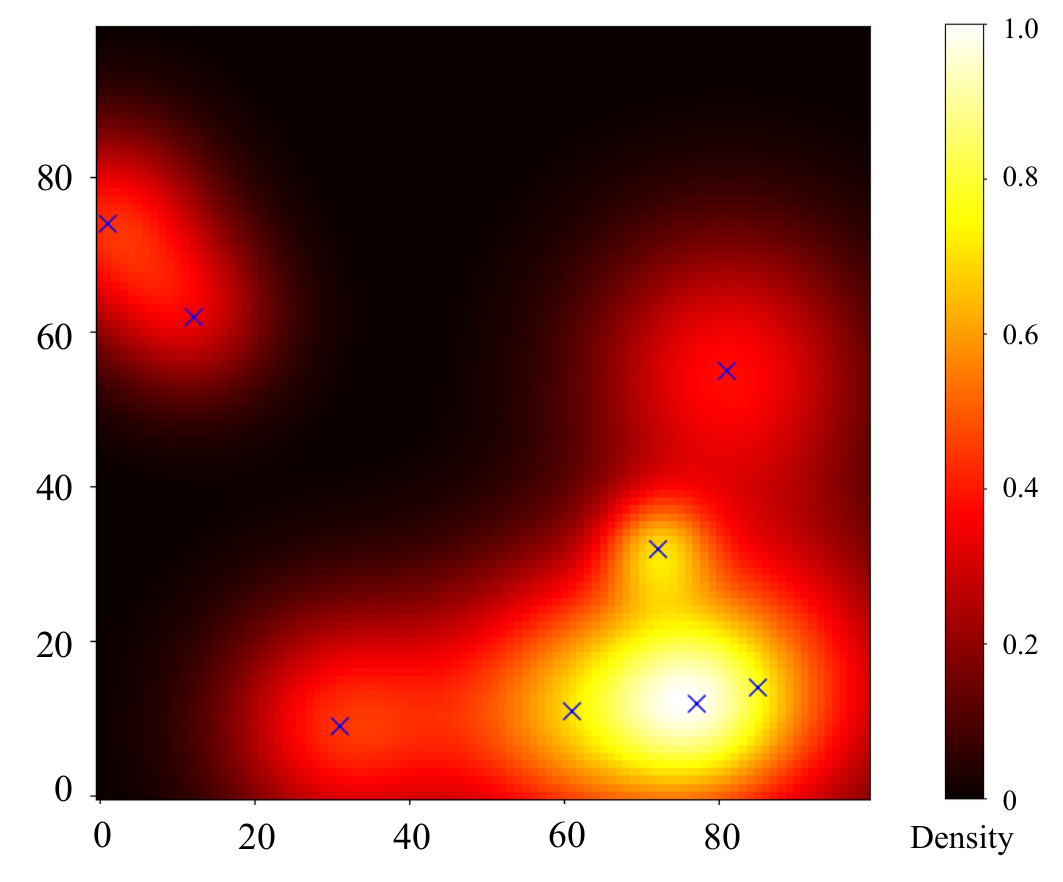}
    \caption{Density map with Hotspots.}
    \label{fig:3}
\end{figure}

To validate the effectiveness of the proposed system, a series of systematic experimental evaluations are conducted, along with representative visualizations of key results. As shown in Fig. \ref{fig:3}, the heatmap illustrates the task density distribution across the urban environment. Red and yellow regions indicate task hotspots, reflecting areas with a higher concentration of task demand, where the system dynamically reallocates UAV resources accordingly.

\begin{figure}[!htbp]
    \centering
    \includegraphics[width=1.0\columnwidth]{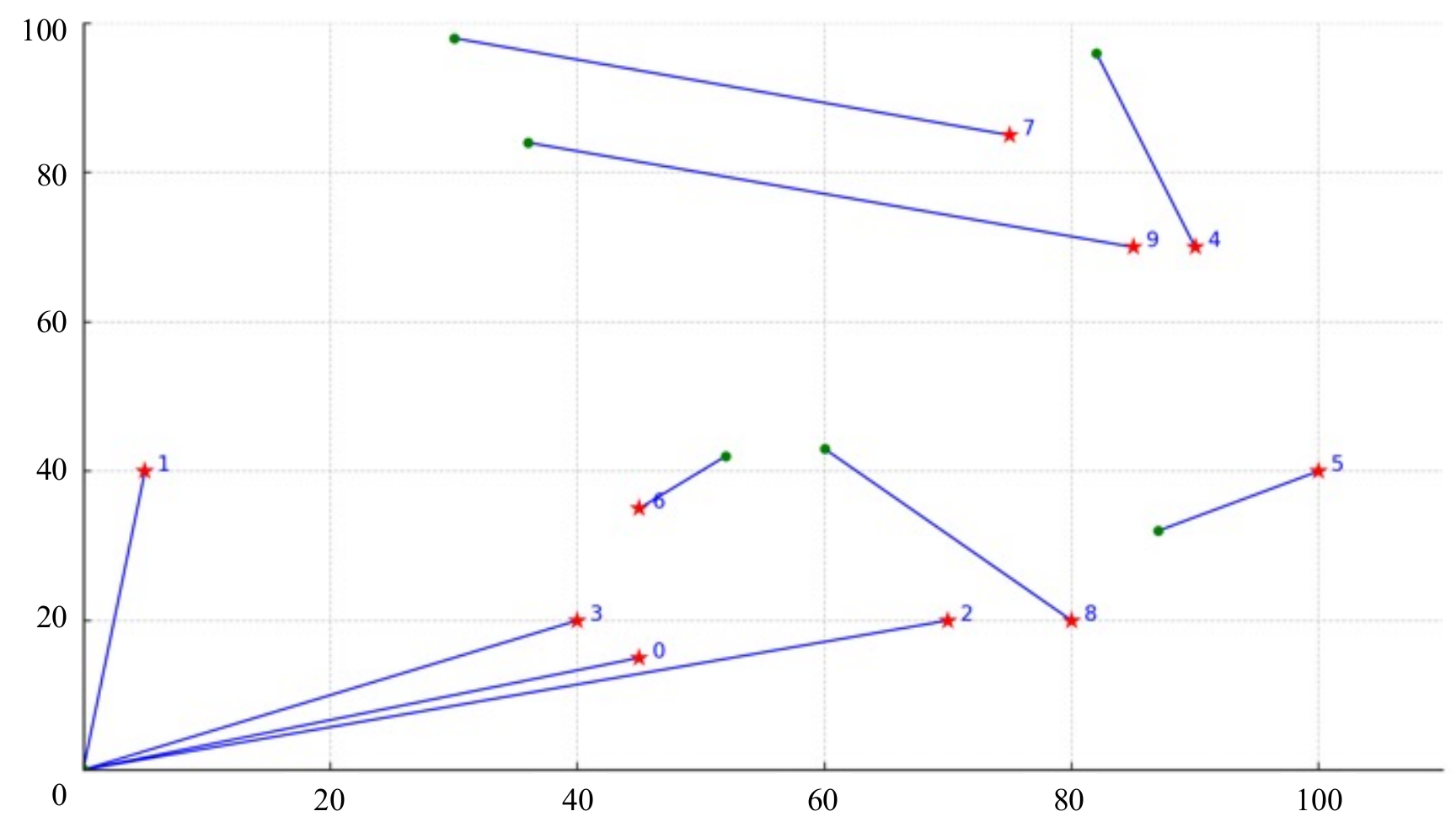}
    \caption{UAV task assignment trajectory route.}
    \label{fig:4}
\end{figure}

Fig. \ref{fig:4} presents the optimized UAV trajectories. Each blue track represents a mission route, with a total of ten, starting from the starting point (green dot) to the task target (red star), showing the global path optimization effect based on the minimum travel distance while maintaining a reasonable spatial distribution to avoid congestion and conflict.

In order to fully verify the performance of the proposed CoordField system, this study designed comparative experiments from two perspectives: on the one hand, comparing the differences in task understanding and execution processes of different LLMs; on the other hand, while ensuring that other parts of the agentic system remain unchanged, further investigating the differences in task performance between different traditional task allocation algorithms and the Coordination field. The following is a detailed analysis.

\begin{table}[h!]
\footnotesize
\centering
\begin{threeparttable}
\caption{Comparison of Task Performance Metrics Across Models}
\begin{tabular}{>{\centering\arraybackslash}p{2.6cm} 
                >{\centering\arraybackslash}p{1cm} 
                >{\centering\arraybackslash}p{1cm} 
                >{\centering\arraybackslash}p{1cm} 
                >{\centering\arraybackslash}p{1cm}}
\toprule
\textbf{Model} & \textbf{TPA} & \textbf{CE} & \textbf{TLB} & \textbf{UUR} \\
\midrule
CoordField & 96\% & 95\% & 0.8 & 97\% \\
Deepseek-v3 & 70\% & 78\% & 2.1 & 85\% \\
GPT-4o & 74\% & 85\% & 1.5 & 82\% \\
Claude-3-7-Sonnet & 76\% & 80\% & 1.6 & 84\% \\
LLaMA-4-Scout & 72\% & 79\% & 1.9 & 83\% \\
Gemini-2.5-Pro & 73\% & 81\% & 1.7 & 84\% \\
\bottomrule
\end{tabular}
\begin{tablenotes}
\footnotesize

\item Model versions: Deepseek-v3 (25.03.24), GPT-4o (24.05.13), Claude-3-7-Sonnet (25.02.19), LLaMA-4-Scout (25.04.05), Gemini-2.5-Pro-Preview (25.03.25).

TPA: Task Parsing Accuracy; CE: Coverage Efficiency; TLB: Task Load Balance; UUR: UAV Utilization Rate.
\end{tablenotes}
\label{tab1}
\end{threeparttable}
\end{table}

The experiment in Table~\ref{tab1} compares the performance of the CoordField system with the current mainstream large language models, in natural language instruction parsing and task collaborative execution. The experimental results show that CoordField has achieved leading performance in all indicators from four dimensions. The multi-agent system designed by the CoordField system can gain advantages in language understanding, multi-task parsing, and resource scheduling, especially in multi-round instruction scenarios, showing stronger generalization and robustness.

\begin{table}[h!]
\footnotesize
\centering
\begin{threeparttable}
\caption{Comparison of performance with traditional algorithms}
\begin{tabular}{>{\centering\arraybackslash}p{2.6cm} 
                >{\centering\arraybackslash}p{1cm} 
                >{\centering\arraybackslash}p{1cm} 
                >{\centering\arraybackslash}p{1cm} 
                >{\centering\arraybackslash}p{1cm}}
\toprule
\textbf{Algorithm} & \textbf{CR} & \textbf{CE} & \textbf{TLB} & \textbf{UUR} \\
\midrule
CoordField         & 89\% & 95\% & 0.8 & 97\% \\
ACO                & 93\% & 84\% & 1.3 & 68\% \\
GWO                & 91\% & 89\% & 1.7 & 75\% \\
WOA                & 97\% & 90\% & 1.7 & 91\% \\
A*                 & 93\% & 75\% & 3.6 & 53\% \\
\bottomrule
\end{tabular}
\begin{tablenotes}
\footnotesize
\item ACO: Ant Colony Optimization; GWO: Grey Wolf Optimizer; WOA: Whale Optimization Algorithm; A\*: A Star Algorithm.

CR: Completion Rate; CE: Coverage Efficiency; TLB: Task Load Balance; UUR: UAV Utilization Rate.
\end{tablenotes}
\label{tab2}
\end{threeparttable}
\end{table}

The comparative experiment in Table~\ref{tab2} evaluates the performance of the Coordination field method by replacing the Coordination field part of the system with four other classic optimization algorithms while ensuring that the entire Agentic system structure remains unchanged. From the experimental data, it can be seen that although the completion rate is slightly better than CoordField, reflecting the accuracy of the traditional optimization algorithm, CoordField performs best in terms of task coverage efficiency and load balance. In contrast, the traditional optimization method has large fluctuations in balance. This group of experiments verifies that the Coordination field method introduced by CoordField can achieve a more efficient multi-UAV collaborative allocation strategy.

\section{conclusion}

This paper proposes a coordination field Agentic system for UAV task coordination. The system demonstrates high efficiency, stability and adaptability, and is able to parse natural language instructions in real time and achieve efficient task allocation through coordination field guidance. Experimental results verify its superior performance in complex and dynamic urban environments. In future work, we plan to extend the system to three-dimensional scenarios and real-world UAV swarms.

\bibliographystyle{IEEEtran}
\bibliography{egbib}

\end{document}